%% file: main.tex
\pdfoutput=1
\documentclass[sigconf]{acmart}

\AtBeginDocument{%
  \providecommand\BibTeX{{%
    \normalfont B\kern-0.5em{\scshape i\kern-0.25em b}\kern-0.8em\TeX}}}

\copyrightyear{2024}
\acmYear{2024}
\setcopyright{rightsretained}
\acmConference[KDD '24]{Proceedings of the 30th ACM SIGKDD Conference on Knowledge Discovery and Data Mining}{August 25--29, 2024}{Barcelona, Spain}
\acmBooktitle{Proceedings of the 30th ACM SIGKDD Conference on Knowledge Discovery and Data Mining (KDD '24), August 25--29, 2024, Barcelona, Spain}\acmDOI{10.1145/3637528.3671632}
\acmISBN{979-8-4007-0490-1/24/08}

\makeatletter
\gdef\@copyrightpermission{
  \begin{minipage}{0.3\columnwidth}
   \href{https://creativecommons.org/licenses/by/4.0/}{\includegraphics[width=0.90\textwidth]{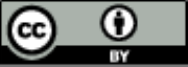}}
  \end{minipage}\hfill
  \begin{minipage}{0.7\columnwidth}
   \href{https://creativecommons.org/licenses/by/4.0/}{This work is licensed under a Creative Commons Attribution International 4.0 License.}
  \end{minipage}
  \vspace{5pt}
}
\makeatother

\usepackage[english]{babel}
\usepackage[utf8]{inputenc} 
\usepackage[T1]{fontenc}    
\usepackage{hyperref}       
\usepackage{booktabs}       
\usepackage{amsfonts}       
\usepackage{nicefrac}       
\usepackage{microtype}      
\usepackage{xcolor}         
\usepackage{multicol}

\usepackage{amsmath,wasysym}
\usepackage[ruled,vlined,linesnumbered, resetcount]{algorithm2e}

\usepackage{subcaption}
\usepackage{multirow}
\usepackage{textcomp}
\usepackage{footnote}
\usepackage{makecell}
\usepackage{xspace}
\usepackage{enumitem}
\usepackage{wrapfig}
\usepackage{amsthm}
\usepackage{wrapfig}


\input{macros}

\title{Large Scale Hierarchical Industrial Demand Time-Series Forecasting incorporating Sparsity}
\author{%
  Harshavardhan Kamarthi}
  \affiliation{
  Georgia Institute of Technology\country{USA}}
  \email{hkamarthi3@gatech.edu}
  \author{%
  Aditya B. Sasanur}
  \affiliation{
  Georgia Institute of Technology\country{USA}}
  \email{asasanur@gatech.edu}
  \author{%
  Xinjie Tong}
  \affiliation{
  The Dow Chemical Company
  \country{USA}}
  \email{xtong1@dow.com}
  \author{%
  Xingyu Zhou}
  \affiliation{
  The Dow Chemical Company
  \country{USA}}
  \email{xzhou14@dow.com}
  \author{%
  James Peters}
  \affiliation{
  The Dow Chemical Company
  \country{USA}}
  \email{japeters@dow.com}
  \author{%
  Joe Czyzyk}
  \affiliation{
  The Dow Chemical Company
  \country{USA}}
  \email{jczyczyk@dow.com}
  \author{%
  B. Aditya Prakash}
  \affiliation{
  Georgia Institute of Technology\country{USA}}
  \email{badityap@cc.gatech.edu}

\renewcommand{\shortauthors}{Harshavardhan Kamarthi et al.}

\begin{document}

\input{Text/abstract}
\maketitle

\input{Text/intro}
\input{Text/related}
\input{Text/problem}
\input{Text/methodology}
\input{Text/experiments}

\input{Text/results}

\input{Text/casestudy}
\input{Text/conclusion}

\bibliographystyle{ACM-Reference-Format}
\bibliography{references}



\end{document}

%% file: macros.tex
\usepackage{xcolor}

\newcommand{\model}{\textsc{HAILS}\xspace}
\newcommand{\profhit}{\textsc{ProfHiT}\xspace}

\newtheorem{theorem}{Theorem}

\newtheorem{definition}{Definition}
\newcommand{\hide}[1]{}

\newcommand{\deepvar}{\textsc{DeepAR}\xspace}
\newcommand{\hiere}{\textsc{HierE2E}\xspace}
\newcommand{\mint}{\textsc{MinT}\xspace}
\newcommand{\erm}{\textsc{ERM}\xspace}

\newcommand{\pembu}{\textsc{PEMBU}\xspace}
\newcommand{\sharq}{\textsc{SHARQ}\xspace}
\newcommand{\mbest}{\textsc{M5-Leader}\xspace}

\def\basemu{{\mu}}
\def\basesigma{{\sigma}}
\def\baselambda{{\lambda}}
\def\refinedmu{{\hat{\mu}}}

\def\refinedsigma{{\hat{\sigma}}}
\def\sparseset{{\mathbb{S}}}
\def\loss{{\mathcal{L}}}

%% file: Text/abstract.tex
\begin{abstract}
Hierarchical time-series forecasting (HTSF) is an important problem for many real-world business applications where the goal is to simultaneously forecast multiple time-series that are related to each other via a hierarchical relation.
Recent works, however, do not address two important challenges that are typically observed in many demand forecasting applications at large companies.
First, many time-series at lower levels of the hierarchy have high sparsity i.e., they have a significant number of zeros. Most HTSF methods do not address this varying sparsity across the hierarchy.
Further, they do not scale well to the large size of the real-world hierarchy typically unseen in benchmarks used in literature.
We resolve both these challenges by proposing \model, a novel probabilistic hierarchical model that enables accurate and calibrated probabilistic forecasts across the hierarchy by adaptively modeling sparse and dense time-series with different distributional assumptions and reconciling them to adhere to hierarchical constraints.
We show the scalability and effectiveness of our methods by evaluating them against real-world demand forecasting datasets. We deploy \model at a large chemical manufacturing company for a product demand forecasting application with over ten thousand products and observe a significant 8.5\% improvement in forecast accuracy and 23\% better improvement for sparse time-series. The enhanced accuracy and scalability make \model a valuable tool for improved business planning and customer experience. 
\end{abstract}

%% file: Text/intro.tex
\section{Introduction}

 Hierarchical time-series forecasting is a problem that profoundly influences decision-making across various domains. These time-series data possess inherent hierarchical relationships and structures~\cite{athanasopoulos2020hierarchical,hyndman2018forecasting}. Instances of such situations include predicting employment trends \citep{taieb2017coherent}
  across diverse geographical scales, forecasting the spread of epidemics \citep{reich2019collaborative}, etc. 
 When dealing with time-series datasets that exhibit underlying hierarchical dependencies, the objective of hierarchical time-series forecasting is to generate precise forecasts for all individual time-series while capitalizing on the hierarchical interconnections among them \citep{hyndman2011optimal}. 
 For instance, at a large manufacturing company, forecasting demand at various levels of aggregation is important~\cite{bose2017probabilistic}.  Forecasts at a middle level of the business hierarchy are important for procuring raw materials and determining the amount of intermediate materials (or product families) to produce in the medium-term.  
 Near-term forecasts at lower levels of the hierarchy relate more to specific products and even package sizes that are needed.  
 Additionally, companies do not forecast based solely using historical data but include external variables (such as macroeconomic forecasts which incorporate reasonable assumptions about the future) to improve demand forecasts.  

Previous forecasting methods have not typically placed an emphasis on providing well-calibrated probabilistic forecasts that model uncertainty. 
Instead, traditional methods have primarily concentrated on providing single-point predictions.
In contrast, recent post-processing techniques \cite{wickramasuriya2019optimal,ben2019regularized,taieb2017coherent}
refine forecast distributions generated by independent base models as a preprocessing step.
These post-processing methods offer the advantage of being readily applicable to forecasts generated by various models and are usually simple to implement and tractable even for large-scale datasets with thousands of time-series in the hierarchy.
However, they fall short in enabling the base forecasting models to grasp the intricate hierarchical relationships among time-series data within the hierarchy.

In contrast, end-to-end learning neural methods have taken a more direct approach by incorporating hierarchical relationships as an integral part of either the model architecture ~\cite{rangapuram2021end} or learning algorithm~\cite{han2021simultaneously}.
These comprehensive end-to-end approaches tend to surpass post-processing methods by imposing hierarchical constraints on forecast distribution parameters, such as the mean or fixed quantiles.
Most end-to-end methods do not consistently enforce hierarchical coherence across the entirety of the distribution.
Some recent methods~\cite{han2021simultaneously} do impose some distributional constraints such as across specific quantiles.
\profhit~\cite{kamarthi2023rigidity} is capable of generating well-calibrated forecasts by imposing hierarchical constraints on the forecast distributions.
Further, large scale industrial demand time-series exhibit a range of distributional behavior across
the hierarchy~\cite{turkmen2021forecasting}.
Importantly, many time-series corresponding to individual products for specific customers at lower levels have high sparsity due to infrequent demand~\cite{jati2023hierarchical,makridakis2022m5}.
This can be attributed to various factors, such as the seasonality, novelty, or niche appeal of products at these levels.
However, time-series at higher levels show much less sparsity being an aggregation of multiple time-series at lower levels.
Post-processing methods, due to their inability to transfer information across base forecasts
cannot capture this wide range of behavior and overcome this just by reconciliation at post-processing.
These methods are not designed to model time-series of different sparsity simultaneously and
also provide subpar performance.

\hide{
\begin{figure}[h]
    \centering
    \includegraphics[width=.9\linewidth]{Images/Figure1.png}
    \caption{\model adapts to large hierarchy of demand time-series with varying sparsity when reconciliation of hierarchical forecasts.}
    \label{fig:firstfig}
\end{figure}
}

Motivated by a real-world use-case at a large scale chemical company,
 we propose \model (Hierarchical forecasting with Adaptation for Industrial and Large Sparse time-series), a novel hierarchical forecasting framework that is both scalable and capable of generating well-calibrated forecasts with precise uncertainty measurements.
We overcome this challenge by proposing to model the lower-level forecasts and higher-level forecasts using appropriate distributions.
At levels of the hierarchy that have a mixture of both distributions,
we use distributional approximations to have uniform distributions across subtree when reconciling the forecasts.
To accomplish this, we propose a novel loss function that enables the model to adapt to sparse time-series data at lower levels of the hierarchy while being able to reconcile with denser time-series at higher levels. 
We summarize our contributions as follows:
\begin{itemize}
    \item \textbf{Efficient Large Scale Probabilistic Hierarchical Forecasting}: We propose a novel hierarchical forecasting framework that is both scalable and capable of generating well-calibrated forecasts with reliable uncertainty measurements for large industrial time-series.
    \item \textbf{Adaptation to Time-series of different levels of sparsity}: We propose modelling the lower level forecasts and higher level forecasts using different distributions based on historical sparsity and propose a novel framework to reconcile them.
    \item \textbf{State-of-art performance on large datasets}: We demonstrate the effectiveness of our proposed method on large-scale demand datasets with thousands of time-series in the hierarchy.
          We evaluate on a public dataset as well as a proprietary dataset from a large chemical manufacturing company.
          We show that our method outperforms the state-of-art methods across most levels of the hierarchy both in terms of accuracy of point forecasts and probabilistic forecasts.
          We perform a detailed case study to demonstrate the impact of our proposed method on a real-world application at a large chemical company.
\end{itemize}

%% file: Text/related.tex
\section{Related Works}
\label{sec:related}

Classical methods in hierarchical time-series forecasting traditionally employed a two-phase method, concentrating on point predictions~\cite{hyndman2011optimal,hyndman2018forecasting}.
These methods predicted time-series at a singular hierarchy level, then extrapolated forecasts to other levels using hierarchical relationships.
Recent techniques, such as \mint and \erm, act as post-processing procedures, refining forecasts across all hierarchy levels. \mint \cite{wickramasuriya2019optimal, wickramasuriya2021probabilistic} operates under the assumption that baseline forecasts are independent and unbiased, aiming to minimize forecast error variance from historical data.
\erm\cite{ben2019regularized} modifies this by not assuming unbiased forecasts.

Recent neural network approaches offer more end-to-end learning of patterns of individual time-series as well as hierarchical relations across time-series. \citet{rangapuram2021end} adopt a  strategy of projecting the forecasts into a subspace of reconciled forecasts via a differentiable operation and optimize the loss on the projected forecasts. \sharq \cite{han2021simultaneously} represents another novel deep-learning probabilistic method that employs quantile regression and regularizes consistency across various forecast distribution quantiles.
\profhit \cite{kamarthi2023rigidity} imposes hierarchical constraints on the forecast distributions by minimizing distributional distance between the parent forecast and the sum of the child forecasts.
However, none of the methods are designed to adapt to sparse time-series and therefore perform sub-optimally on real-world industrial demand time-series.

%% file: Text/problem.tex
\section{Problem Statement}
We denote the dataset $\mathcal{D}$ of $N$ time-series over the time horizon $1,2,\dots,T$. Let $\mathbf{y}_i \in \mathrm{R}^{T}$ be time-series $i$ and $y_{i}^{(t)}$ its value at time $t$. The hierarchical relations across time-series is denote as $\mathcal{T} = (G_{\mathcal{T}}, H_{\mathcal{T}})$ where $G_{\mathcal{T}}$ is a tree of $N$ nodes rooted at time-series $1$ (time-series $1$ is the aggregate of all leaf time-series). 
Consider a non-leaf node (time-series) $i$ with children $\mathcal{C}_i$. 
The hierarchical relations are of the form
$H_{\mathcal{T}} = \{ \mathbf{y}_{i} = \sum_{j \in \mathcal{C}_i} \phi_{ij}\mathbf{y}_{j}: \forall i \in \{1, 2, \dots, N\}, |\mathcal{C}_i|  > 0 \}$
where values of $\phi_{ij}$ are constant and known. 

Our problem can be formulated as follows: Given a dataset $\mathcal{D}$ with underlying hierarchical relations $H_\mathcal{T}$, we learn a model $M$ that provides \emph{{accurate}} probabilistic forecast distributions \\$\{p_M(y_1^{(t+1)}|\mathcal{D}^t),\dots p_M(y_N^{(t+\tau)}|\mathcal{D}^t)\}$ across all levels of the hierarchy where $\tau$ is the forecast horizon.

%% file: Text/methodology.tex
\section{Methodology}

\begin{figure*}[!thb]
    \centering
    \includegraphics[width=.9\linewidth]{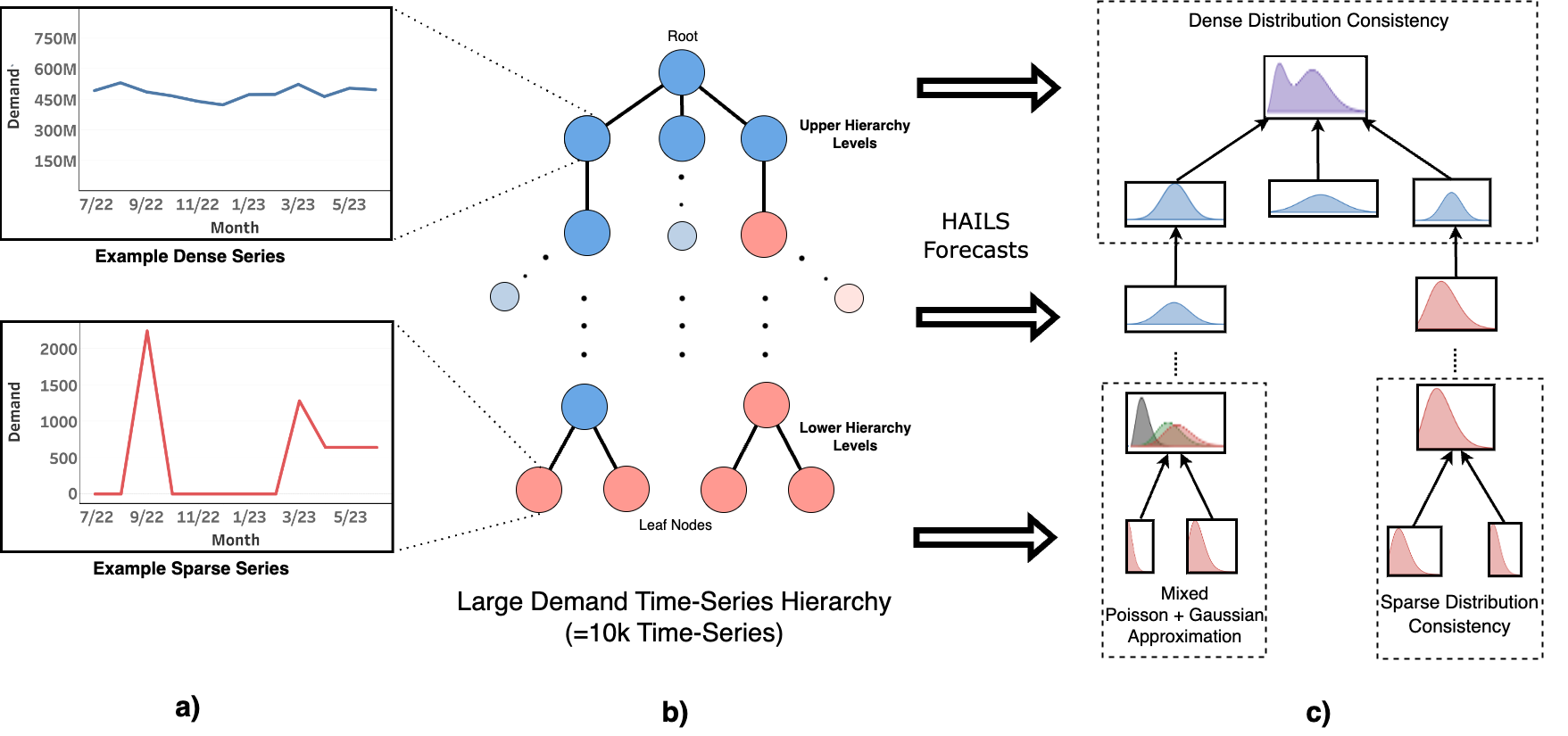}
    \caption{Overview of pipeline of \model.
    (a) The lower levels of the hierarchy tend to have {\color{red} sparse} (red) time-series while the higher levels have {\color{blue}denser} (blue) time-series.
    (b) \model first generates forecasts for each of the time-series of hierarchy with their parametric form depending on the sparsity of the time-series. The denser time-series forecasts are modeled as Gaussians and sparser ones as Poisson. 
    (c) The distributions are reconciled via a distribution consistency loss for each subtree. If the subtree has all distribution same the appropriate loss is applied. In case of mixed subtrees, Poisson distribution of children are first approximated as Gaussian.
    }
    \label{fig:main}
\end{figure*}

Most hierarchical forecasting models
struggle to adapt to large hierarchies found in real-world industrial applications both in terms of effectiveness and efficiency 
in learning from tens of thousands of time-series as well as adapting to sparse time-series at the lower levels of the hierarchy.
\model overcomes these challenges in two ways. 
First, we make architectural design choices to enable more efficient in learning from larger hierarchies as well as support sparse time-series.
Second, we develop optimization methods to allow for learning accurate and consistent forecasts both for sparse and dense forecasts
at different levels of the hierarchy.
We first provide a brief overview of \profhit, state-of-art hierarchical forecasting model and then discuss in detail our innovations
to enable dealing with these challenges.

\subsection{Probabilistic Hierarchical Forecasting}
\profhit~\cite{kamarthi2023rigidity} is a state of the art probabilistic forecasting model that.
It optimizes the full distribution of forecasts of the hierarchy to be both accurate and consistent with the hierarchical constraints.
It first produces base forecasts for each node of the hierarchy
independently via a differentiable neural model.
The authors of \profhit chose to use CaMuL~\cite{kamarthi2021camul},
a state of the art neural probabilistic forecasting model to produce
the base forecasts parameterized by normal distribution $\{(\basemu_i, \basesigma_i)\}_{i=1}^N$.
The base forecast parameters are used as prior distribution parameters to generate refined distribution that leverage inter-series relations and hierarchical constraints to produce the refined parameters $\{(\basemu_i, \basesigma_i)\}_{i=1}^N$.
This is achieved by the \textit{Hierarchy-aware Refinement Module}
and the whole model is trained  on both the Log-likelihood loss for
\textit{accuracy} and Soft Distributional Consistency Regularization
(SDCR) for \textit{Distributional Consistency} by minimizing the Distributional Consistency Error defined as follows:
\begin{definition}(Distributional Consistency Error)~\cite{kamarthi2023rigidity}
    Given the forecasts at time $t+\tau$ as $\{p_M(y_1^{(t+\tau)}|\mathcal{D}^t),\dots p_M(y_N^{(t+\tau)}|\mathcal{D}^t)\}$ distributional consistency error (DCE) is defined as
        {
            \begin{equation}
                \sum_{i\in \{1,\dots,N\}, \mathcal{C}_i\neq \emptyset} Dist\left(p_M(y_i^{(t+\tau)}|\mathcal{D}^t), p_M(\sum_{j\in \mathcal{C}_i} \phi_{i,j} y_j^{(t+\tau)}|\mathcal{D}^t) \right)
                \label{eqn:dce}
            \end{equation}
        }
    where $Dist$ is a distributional distance metric.
\end{definition}
The final forecasts $\{P(y_{i}^{(t+\tau)}|\mathcal{D}^t)\}_{i=1}^N$
are thus optimized to be accurate and distributionally consistent across the hierarchy.

\subsection{\model: Forecasting for large hierarchies with sparse time-series}
\model models sparse and dense time-series using appropriate distributions when forecasting.
A key challenge which we overcome in the process is to 
enable learning consistent forecasts in cases where parts of the hierarchy have both sparse and dense time-series.
To enable this feature, \model proposes important architectural changes to \profhit
and a novel loss: \textit{Distributional consistency regularization with Sparse adaptation}.
We describe the various modules of \model as follows.

\subsubsection{Testing for Poisson Distribution}
We first determine whether we should model a given node of the hierarchy as a sparse time-series. 
We use the Poisson distribution to model the time-series if it is deemed sparse since we can model the high probability of observing zeros.
Therefore, to systematically classify the data from a given node of the hierarchy as sparse, we use the \textit{Poisson dispersion test} on samples from training data of the time-series.
Intuitively, the dispersion test tests if the mean and variance of the data samples are similar.
We observe that using a $p$ value threshold of 0.1 is a good measure to classify nodes as \textit{sparse} or \textit{dense}.
We also make sure that the parents of a node classified as dense are automatically dense.
We observe this to be always true for our benchmark datasets.
But, in case it does not hold, we explicitly classify the parents as dense.
Notationally, we denote all nodes in $\{1, \dots, N\}$ that are classified as sparse as $\mathbb{S}$.

\subsubsection{Base forecasting model}
The requirements for choosing a base forecasting model are based on the application's specific
needs. 
\profhit uses CaMuL ~\cite{kamarthi2021camul,kamarthi2021doubt}
due to its superior performance in terms of accuracy and uncertainty
quantification.
However, it makes deploying to large industrial hierarchies infeasible.
First, CaMuL is a stochastic model~\cite{louizos2019functional} that leverages multiple sampling components that makes stable training a hard technical challenge
when scaling it to train tens of thousands of time-series independently.
Secondly, it requires significant amount of historical data that it uses as \textit{reference points} to map similar patterns from historical data to current input time-series for uncertainty quantification.
Along with the challenge of the high compute requirement of storing and embedding
these historical time-series, in many real-world applications
we do not have sufficient historical data to learn reliably.
Finally, CaMuL is not designed to model sparse predictions and instead parameterized the output as a Gaussian.
Instead, we chose
a simpler model: a Gated Recurrent Unit (GRU) neural network, a widely adopted
recurrent deep learning model.
Depending on the nature of the node, the base forecasts output the
forecast parameter.
For a node $i \not\in \sparseset$ classified as dense, it outputs two parameters of the normal distribution: $(\basemu_i, \exp(\basesigma_i))$.
For any node $j \in \sparseset$ classified as sparse, it simply outputs only the Poisson mean parameter: $\baselambda_j = \basemu_j$.

\subsubsection{Hierarchy-aware Refinement Module}

This module uses the base forecasts from RNNs and refined them to 1) leverage information of the time-series across the hierarchy 2) enables them to be distributionally consistent by training on the SDCR.
Let $\mathbf{\basemu} = [\basemu_{1} \dots, \basemu_N]$ be a vector of
means of base distributions for all nodes.
$\refinedmu_i$ is the weighted sum of $\basemu_i$ and base mean of all time-series:
{\begin{equation}
    \gamma_i = \text{sigmoid}(\hat{w}_i), \quad
    \refinedmu_i = \gamma_i \basemu_i + (1-\gamma_i) \mathbf{w}_i^T\mathbf{\basemu}
    \label{eqn:corem1}
\end{equation}}
where $\{\hat{w}_i\}_{i=1}^N$ and $\{\mathbf{w}_{i}\}_{i=1:N}$ are parameters of the model and $\text{sigmoid}(\cdot)$ denotes the sigmoid function.
$\gamma$ intuitively denotes the tradeoff between relying on the base
mean and information from rest of the distribution.
Let $\mathbf{\basesigma} = \{\basesigma_{i}| i \not\in \sparseset\}$ 
be a vector of variances for dense nodes' base forecasts.
The variance parameter $\refinedsigma_i$ of the refined distribution is derived from the base distribution parameters 
{\begin{equation}
            \refinedsigma_i = c\basesigma_i \text{sigmoid}(\mathbf{v}_{1i}^T\mathbf{\basemu} + \mathbf{v}_{2i}^T \mathbf{\basesigma} + b_i)
            \label{eqn:corem2}
        \end{equation}}
where $\{\mathbf{v}_{1i}\}_{i=1}^N$, $\{\mathbf{v}_{2i}\}_{i=1}^N$ and $\{b_i\}_{i=1}^N$ are parameters and $c$ is a positive constant hyperparameter.

\begin{figure*}[!thb]
    \centering
    \includegraphics[width=.9\linewidth]{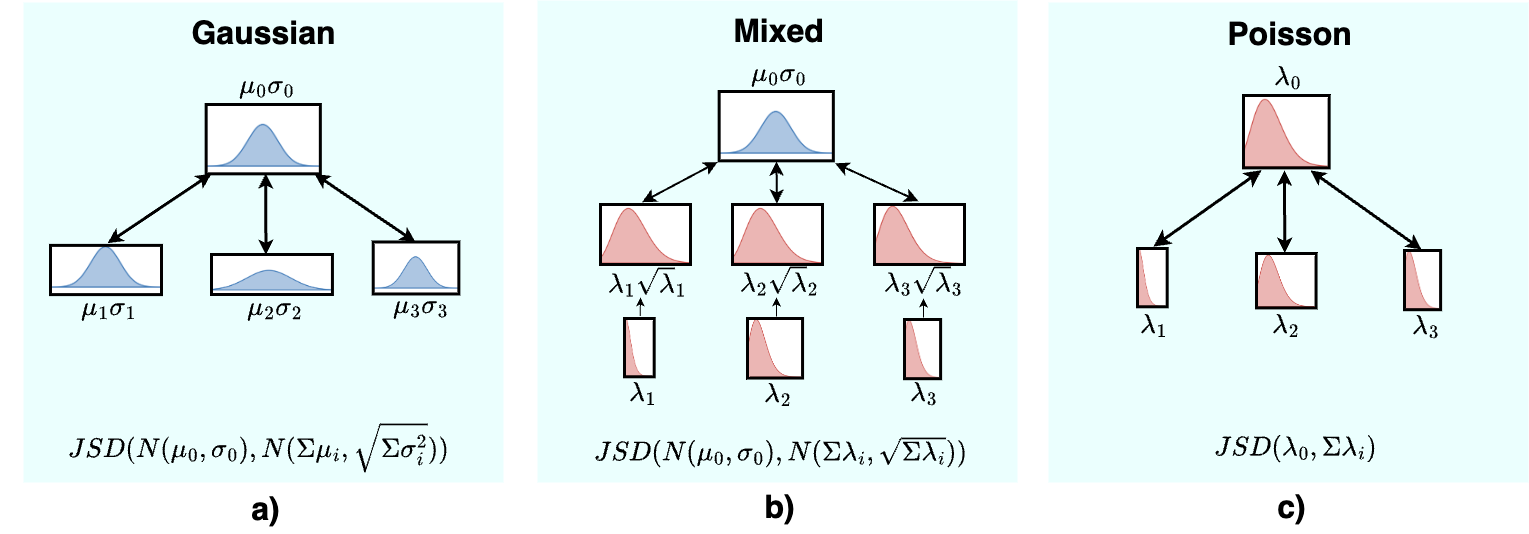}
    \caption{For homogeneous subtrees of Normal and Poisson distribution (a,c), JSD divergence loss is applied directly. In case of heterogeneous subtrees (b), Poisson distributions are first approximated as gaussians and then JSD is applied across resultant gaussians.
    }
    \label{fig:adaptive}
\end{figure*}

\subsubsection{Soft Distributional Consistency Regularization}

\profhit learns to generate forecasts that are distributionally consistent by introducing SDCR.
It forces the model to minimize the Distributional Consistency Error
across the forecasts of the hierarchy leading to the aggregated forecasts of the children being similar to the parent forecast.
However, SDCR only deals with dense time-series since it models them as Gaussians. Moreover. it cannot deal with different types 
 of distributions across the hierarchy. 
 Therefore,
SDCR cannot be directly applied to hierarchies that have sparse time-series ($\sparseset \neq \Phi)$.
\model introduces \textit{Distributional Consistency
Regularization with Sparse adaptation} (DCRS) that allows for
hierarchies with varying time-series sparsities to  provide
distributional consistency.
DCRS applies different consistency losses across the subtrees of the
hierarchies based on the the sparsity of the parents and children.
We specifically look at three cases that are observed in the hierarchies:

\noindent\textit{Dense Parent-Dense Children:} If the parent node
as well as children nodes are dense, we use the same distributional
consistency loss as \profhit:
we model the parent and children forecast distributions
as gaussians and compute the Jenson-Shannon divergence:
{\begin{equation}
    \begin{split}
        \loss_{DCRS}^{(i)} = JSD(\mathcal{N}(\refinedmu_i, \refinedsigma_i)&|\mathcal{N}\left(\sum_{j\in C_i}\phi_{ij}\refinedmu_j, \sqrt{\sum_{j\in C_i}\phi_{ij}^2\refinedsigma^2_j}\right)) = \\
         \sum_{i=1}^N \frac{\refinedsigma_i^2 + \left( \refinedmu_i - \sum_{j\in C_i} \phi_{ij}\refinedmu_j \right)^2}{4\sum_{j\in C_i}\phi^2_{ij} \refinedsigma_j^2} +
        &\sum_{i=1}^N  \frac{\sum_{j\in C_i}\phi^2_{ij} \refinedsigma_j^2 + \left( \refinedmu_i - \sum_{j\in C_i} \phi_{ij}\refinedmu_j \right)^2}{4\refinedsigma_i^2} 
    \end{split}
    \label{eqn:jsd}
\end{equation}}

\noindent\textit{Sparse Parent- Sparse Children:}
To calculate the distributional consistency error of sparse time-series at lower levels of the hierarchy we note that we assume the forecasts are Poisson distributions. The JSD between two Poissons has
a closed form solution:
\begin{equation}
\begin{split}
   \loss_{DCRS}^{(i)}&= JSD\left(\lambda_i| \sum_{j\in C_i}\lambda_j\right) = \lambda_i \log\left(\frac{\lambda_1}{\sum_{j\in C_i}\lambda_j}\right)\\
    &+ \sum_{j\in C_i}\lambda_j \log\left(\frac{\sum_{j\in C_i}\lambda_j}{\lambda_i}\right).
    \end{split}
\end{equation}

\noindent\textit{Mixed Subtrees:}
Now we examine the case where the parent is a dense node but some
or all of her children are sparse.
We note that as we go further up the hierarchy, sparsity of time-series decreases. Therefore, the sparsity assumption on these nodes gets weaker.
We therefore propose to approximate the sparse forecasts of these time-series as Gaussian distributions and apply Eq. \ref{eqn:jsd}
to optimize for distributional consistency.
We perform the approximation leveraging the central limit theorem as follows:
\begin{theorem}
    Let $X_1, X_2, \dots, X_N$ be $N$ independent Poisson random variables with
    parameters $\lambda_1, \lambda_2, \dots, \lambda_N$. 
    Then denote $Y$ as $Y=\sum_{i=1}^N X_i$. Then $Y$ is a Poisson variable with
    parameter $\lambda_Y = \sum_{i=1}\lambda_i$.
    Then for sufficiently large $\lambda_Y$, $Y$ can be approximated by a
    Gaussian distribution $\Tilde{Y} = \mathcal{N}(\lambda, \sqrt{\lambda})$~\cite{lesch2009some}.
\end{theorem}
Therefore, all the children time-series forecasts of sparse node $j$ of form $\lambda_j$ are converted to normal distribution $\mathcal{N}(\lambda_j, \sqrt{\lambda})$. 
Then, once all the forecast distributions are modeled as Normal distributions, we apply Eq.~\ref{eqn:jsd}.

We summarize the three cases of performing distributional consistency in Figure \ref{fig:adaptive}.
The total DCRS loss is denoted as:
\begin{equation}
    \loss_{DCRS} = \sum_{i: |C_i| > 0} \loss_{DCRS}^{(i)}
\end{equation}

\subsubsection{Other Training details}

\noindent\textit{Likelihood loss.}
Similar to \profhit, we use a Log-likelihood loss along with the Distributional coherency loss over all the nodes of the hierarchy.
The log-likelihood loss $\loss_{LL}$ is trained on the refined parameters of the
forecast distribution and summed across all nodes.
The total loss is $\loss = \loss_{LL} + \gamma \loss_{DCRS}$
where the hyperparameter $\gamma$ dictates the relative importance of the
importance of DCRS loss.

\noindent\textit{Pre-train base RNN models:} Pre-training the weights of some of the layers of a neural model to improve the overall efficiency and convergence of a model is a well-known effective technique~\cite{hinton2006reducing}.
We therefore, first train the only RNN modules for each node
independently for point forecasting for small (about 50) epochs before we train all the modules of \model for hierarchical probabilistic forecasting.

\noindent\textit{Hyperparameters}
We use a bidirectional GRU with 60 hidden units for all nodes of the time-series.
For each node we use a 80-20 train validation split to tune the hyperparameters.
For training we use batch size of 32 and learning rate of 0.001.
We use early stopping to determine number of epochs to train. We observed that \model usually converges within 200 epochs for both datasets.

\noindent\textit{Data Preprocessing}
We first normalize the data as follows:
For each non-leaf time-series we divide the time-series value by number of children. Then we 
use the weights $\phi_{ij} = \frac{1}{|C_i|}$
for hierarchical relations. This is so that the higher levels of the hierarchy do not hoave very large values as inputs ot the model to enable stable training.

%% file: Text/experiments.tex
\input{Tables/M5acc.tex}
\input{Tables/m5CRPS}
\section{Experiments}
\subsection{Setup}
We evaluate \model against top hierarchical forecasting baselines on two large hierarchical demand forecasting benchmarks. 
We evaluated all models on a system with Intel 64-core Xeon Pro-
cessor with 128 GB memory and Nvidia Tesla V100 GPU with 32 GB
VRAM. We provide our implementation of \model at \url{https://github.com/AdityaLab/HAILS}.
We used PyTorch for training neural networks and Numpy for other data processing steps.

\subsubsection{Datasets}
While most hierarchical forecasting benchmarks consist of small hierarchies with all time-series being dense, we
choose to evaluate on two benchmarks for our specific application: large hierarchies for demand forecasting. We choose one public dataset and also evaluate on a proprietary real-world use-case of product demand forecasting at a large chemical company.

\noindent\textbf{M5 dataset:}  M5 forecasting competition featured a monthly retail sales forecasting dataset with hierarchically structured sales data with intermittent and erratic characteristics~\cite{makridakis2022m5,syntetos2005accuracy}. The dataset had 12 levels of hierarchy and consisted of 3914 time-series in total. The forecast horizon was up to 28 months ahead.

\noindent\textbf{Dow Demand forecasting:} 
The dataset contains monthly historical sales (in volume) from January 2018 to June 2023 made by Dow in 10+ major industries across 160+ countries. The dataset has a hierarchical structure where the top levels represents the aggregated sales at the country and industry levels, and the lower levels contain the sales data in more granular product classes. The historical sales from January 2018 to June 2022 along with external business indicators were used to train the model. Product demand forecasts were generated for July 2022 to June 2023, and the actual sales during this period are used as ground truth. The forecast horizon was 12 months ahead with the following hierarchical structure:

\begin{table}[h]
\centering
\begin{tabular}{lcc}
\hline
Level & Time-series & Sparsity \\ \hline
L1 (Area) & 4 & 0\% \\
L2 (Country) & 25 & 0\% \\
L3 (Industry) & 418 & 3.69\% \\
L4 (Business Group) & 1111 & 15.61\% \\
L5 & 1956 & 21.05\% \\
L6 & 3462 & 32.76\% \\
L7 & 5459 & 36.49\% \\
L8 & 7587 & 40.93\% \\ \hline
\end{tabular}
\caption{Number of time-series and sparsity (\% of zeros)  by Level for Dow time-series.}
\label{table:level_time_series_sparsity}
\end{table}

\subsubsection{Baselines}
We compare \model's performance against state-of-the-art hierarchical forecasting methods as well as generic time-series forecasting methods.
We first compare against a standard heuristic of averaging past 6 months' values (6-Average).
ARIMA~\cite{makridakis1997arma} is a commonly used statistical time-series models.
We also use GRU~\cite{chung2014empirical} without any reconciliation as a common neural RNN-based forecasting baseline.
For GRU, we used Monte-Carlo dropout~\cite{gal2016dropout} to generate multiple forecast samples for probabilistic forecasts.
Finally we also considered \deepvar~\cite{salinas2020deepar}, popular deep probabilistic forecasting models which do not exploit hierarchy relations.  Note that 6-Average cannot produce probabilistic forecasts due to its deterministic mechanics.

In the case of hierarchical forecasting, we considered \pembu~\cite{taieb2017coherent}, the state-of-art post-processing method applied on \deepvar forecasts reconciled by \mint.
With respect to the state-of-art neural hierarchical forecasting methods, we compare against \sharq~\cite{han2021simultaneously}
a deep learning-based approach that reconciles forecast distributions by using quantile regressions and making the quantile values consistent.
We also compare against \hiere~\cite{rangapuram2021end}, a deep learning-based approach that projects the base predictions onto a space of consistent forecasts and trains the model in an end-to-end manner.
For the M5 benchmark, we also include the scores from the top submission of the M5 competition, denoted as \mbest.

\subsubsection{Evaluation Metrics}
We evaluate our models and baselines using carefully chosen metrics
to measure both point accuracy and probabilistic distribution calibration of the forecasts.
For a ground truth $y^{(t)}$, let the predicted probability distribution be $\hat{p}_{y^{(t)}}$ with mean $\hat{y}^{(t)}$. Also let $\hat{F}_{y^{(t)}}$ be the CDF.



\noindent$\bullet$\textbf{Weighted Root Mean Squared Scaled Error (WRMSSE)} is a scale-invariant metric for point-predictions that can be used to compare across different time-series of varying scales. RMSSE for a time-series is defined as:
\[RMSSE = \sqrt{\frac{1/N \sum_{t=n}^{n+N} (y^{(t)} - \hat{y^{(t)}})^2}{1/(n-1) \sum_{t=2}^{n} (y^{(t)} - {y^{(t-1)}})^2}}\]
where $N$ is the forecast horizon and $n$ is the length of the training data.
We then weight each of the time-series's RMSSE with the average value of ground truth in training dataset to get the weighted RMSSE.
This metric was used in the M5 competition to evaluate the accuracy of point predictions~\cite{makridakis2022m5}.

\noindent$\bullet$\textbf{Cumulative Ranked Probability Score (CRPS)} is a widely used standard metric for the evaluation of probabilistic forecasts that measures \textit{both accuracy and calibration}.
Given ground truth $y$ and the predicted probability distribution $\hat{p}_y$, let $\hat{F}_y$ be the CDF. Then, CRPS is defined as:
\[CRPS(\hat{F}_y, y) = \int_{-\infty}^\infty (\hat{F}_y(\hat{y}) - \mathbf{1}\{\hat{y}>y\})^2 d\hat{y}.\] We approximate $\hat{F}_y$ as a Gaussian distribution formed from samples of the model to derive CRPS.
We normalize the value of CRPS for each time-series by the average value of ground-truth in training data to get \textit{normalized CRPS}.

\noindent$\bullet$\textbf{Root Mean Squared Error (RMSE)} is used to calculate the forecast performance at specific time-step since the opther metrics are usually used to calculate over the full forecast horizon. \[RMSE = \sqrt{\frac{1}{N} \sum_{i=1}^N (y_i - \hat{y_i})^2}\] where N is the number of observations. 

%% file: Tables/M5acc.tex
\begin{table*}[h]
    \centering
    \caption{Weighted RMSSE for M5 dataset. \model achieves the best (bold) or second-best (underline) performance across most levels of hierarchy.}
    \label{tab:m5rmsse}
    \begin{tabular}{c|c|cccccccccccc}
        Model       & \textbf{Total} & L1    & L2    & L3    & L4    & L5    & L6    & L7    & L8    & L9     & L10   & L11   & L12   \\\hline
        Sparsity    &       & 0.1   & 0.1   & 0.1   & 0.1   & 0.2   & 0.5   & 3.1   & 7.4   & 11.5   & 21.7  & 28.3  & 37.3  \\\hline
        6-Average       & 0.851 & 0.472 & 0.534 & 0.554 & 0.614 & 0.693 & 0.711 & 0.775 & 0.913 & 0.871  & 0.985 & 0.986 & 0.984 \\
        ARIMA       & 0.681 & 0.271 & 0.392 & 0.455 & 0.491 & 0.577 & 0.631 & 0.695 & 0.744 & 0.871  & 0.989 & 0.994 & 0.993 \\
RNN         & 0.653 & 0.231 & 0.337 & 0.274 & 0.375 & 0.413 & 0.533 & 0.581 & 0.572 & 0.766  & 0.968 & 0.984 & 0.997 \\
DeepAR      & 0.612 & 0.216 & 0.342 & 0.316 & 0.322 & 0.384 & 0.481 & 0.529 & 0.618 & 0.669  & 0.873 & 0.996 & 0.965 \\
DeepAR-MinT & 0.592 & \underline{0.201} & 0.317 & 0.301 & 0.328 & 0.356 & 0.432 & 0.504 & 0.628 & 0.674  & 0.819 & 0.959 & 0.997 \\
DeepAR-ERM  & 0.585 & 0.221 & 0.283 & 0.275 & 0.316 & 0.384 & 0.442 & 0.481 & 0.611 & 0.629  & 0.779 & 0.986 & 0.969 \\
HierE2E     & 0.614 & 0.215 & 0.291 & 0.318 & \underline{0.337} & 0.397 & \underline{0.405} & 0.477 & 0.656 & 0.748  & 0.886 & 0.924 & 0.966 \\
        SHARQ       & 0.565 & 0.24  & 0.391 & 0.352 & 0.425 & 0.491 & 0.552 & 0.591 & 0.582 & 0.6864 & 0.991 & 0.994 & 0.981 \\
        PEMBU-MINT  & 0.534 & 0.23  & 0.327 & 0.41  & 0.342 & 0.411 & 0.445 & 0.481 & 0.492 & 0.582  & 0.991 & 0.951 & 0.899 \\
        \mbest  & \underline{0.512} & \textbf{0.199} & 0.31  & 0.422 & \textbf{0.277} & \textbf{0.366} & \textbf{0.39}  & \textbf{0.474} & 0.48  & 0.573  & 0.966 & 0.929 & \underline{0.884} \\
        \profhit    & 0.551 & 0.245 & \textbf{0.216} & \underline{0.316} & 0.337 & 0.417 & 0.432 & 0.474 & \textbf{0.439} & \underline{0.557}  & \underline{0.849} & \underline{0.941} & 0.932 \\\hline
        \model      & \textbf{0.502} & 0.211 & \underline{0.233} &\textbf{ 0.262} & \underline{0.311} & \underline{0.382} & 0.416 & \underline{0.462} & \underline{0.443} & \textbf{0.539}  & \textbf{0.693} & \textbf{0.882} & \textbf{0.814}
    \end{tabular}
\end{table*}

%% file: Tables/m5CRPS.tex
\begin{table*}[h]
    \centering
    \caption{Normalized CRPS for M5 dataset. \model achieves the best performance across all levels of hierarchy. \model achieves the best (bold) or second-best (underline) performance across most levels of hierarchy.}
\label{tab:m5crps}
    \begin{tabular}{c|c|cccccccccccc}
        Model       & \textbf{Total} & L1    & L2    & L3    & L4    & L5    & L6    & L7    & L8    & L9     & L10   & L11   & L12   \\\hline
        Sparsity    &       & 0.1   & 0.1   & 0.1   & 0.1   & 0.2   & 0.5   & 3.1   & 7.4   & 11.5   & 21.7  & 28.3  & 37.3  \\\hline
        ARIMA       & 5.233 & 2.563 & 2.742 & 3.643 & 4.145 & 5.672 & 7.264 & 7.984 & 9.335 & 10.445 & 13.244 & 16.264 & 17.894 \\
RNN         & 0.442 & 0.285 & 0.277 & 0.293 & 0.322 & 0.527 & 0.766 & 0.912 & 0.935 & 0.982  & 0.993  & 0.994  & 1.144  \\
DeepAR      & 0.423 & 0.271 & 0.269 & 0.274 & 0.361 & 0.481 & 0.718 & 0.897 & 0.917 & 0.995  & 0.994  & 0.986  & 1.211  \\
DeepAR-MinT & 0.492 & 0.224 & 0.253 & 0.248 & 0.339 & 0.441 & 0.683 & 0.863 & 0.884 & 0.956  & 0.948  & 0.981  & 0.974  \\
DeepAR-ERM  & 0.229 & 0.214 & 0.231 & 0.283 & 0.318 & 0.429 & 0.668 & 0.822 & 0.826 & 0.926  & 0.956  & 0.977  & 0.993  \\
HierE2E     & 0.126 & 0.113 & 0.111 & \underline{0.117} & 0.309 & 0.392 & 0.592 & 0.731 & \underline{0.718} & 0.885  & 0.933  & 0.942  & 0.942  \\
SHARQ       & 0.139 & 0.082 & 0.283 & 0.294 & 0.323 & 0.388 & 0.732 & 0.782 & 0.811 & 0.895  & 0.926  & 0.993  & 0.942  \\
PEMBU-MINT  & 0.126 & 0.064 & \underline{0.067} & \textbf{0.074} & 0.298 & 0.493 & 0.693 & 0.750 & 0.841 & 0.943  & 0.972  & 0.991  & 0.973  \\
\mbest  & \underline{0.119} & \textbf{0.029} & 0.087 & 0.188 & \textbf{0.173} & \underline{0.283} & \underline{0.429} & \underline{0.572} & \textbf{0.715} & \underline{0.774}  & \underline{0.881}  & \underline{0.893}  & \underline{0.942}  \\
\profhit     & 0.132 & 0.032 & 0.088 & 0.163 & 0.294 & 0.481 & 0.622 & 0.637 & 0.824 & 0.872  & 0.937  & 0.924  & 0.982  \\ \hline
\model   & \textbf{0.081} & \underline{0.030} & \textbf{0.060} & 0.152 & \underline{0.246} & \textbf{0.257 }& \textbf{0.380} & \textbf{0.521} & 0.763 & \textbf{0.717}  & \textbf{0.570}  & \textbf{0.826}  & \textbf{0.728} 
    \end{tabular}
\end{table*}

%% file: Text/results.tex
\subsection{Results}

We evaluate the performance of \model on the M5 dataset and then perform a detailed case study showcasing the impact
of \model on demand forecasting at Dow.

\subsubsection{Forecasting performance on M5}

We evaluate the forecasting performance at each of the individual levels and across the entire hierarchy in Table \ref{tab:m5rmsse}. 
We observe that the average performance of  \model is significantly better than all the other baselines as well as \profhit and \mbest across the hierarchy as well as in most of the hierarchy levels. 
Specifically, we observe significant increase of about 12\% in performance at lower levels (L10- L12) with sparse time-series over \profhit, showcasing the importance of leveraging poisson distributions at the lower level and using the novel DCRS loss.
Overall \model achieves best or close to best performance at all levels of the hierarchy.

In terms of the performance of probabilistic forecasts (Table \ref{tab:m5crps}), we also observe over 40\% better CRPS scores of \model over \profhit and 32\% over best baselines with consistently better performance across all levels of the hierarchy.
Similarly, we observe a significant 20\% better performance at the lower levels of the hierarchy.

\input{Tables/Dowacc.tex}
\input{Tables/DowCRPS}

%% file: Tables/Dowacc.tex
\begin{table*}[h]
    \centering
    \caption{Weighted RMSSE for Dow dataset. \model achieves the best (bold) or second-best (underline) performance across all levels of hierarchy.}
    \label{tab:dowrmsse}
    \begin{tabular}{c|c|cccccccc}
        Model        & \textbf{Total} & L1    & L2    & L3    & L4    & L5    & L6    & L7    & L8    \\\hline
        Sparsity     &       & 0   & 0   & 3.69   & 15.61   & 21.05   & 32.72  & 36.49  & 40.93  \\\hline
        6-Average        & 0.814 & 0.477 & 0.492 & 0.612 & 0.731 & 0.855 & 0.923 & 0.987 & 0.985 \\
        ARIMA        & 0.582 & 0.215 & 0.225 & 0.304 & 0.381 & 0.592 & 0.778 & 0.924 & 0.973 \\
        RNN          & 0.527 & 0.187 & 0.176 & 0.244 & 0.287 & 0.698 & 0.729 & 0.988 & 0.997 \\
DeepAR       & 0.496 & 0.143 & 0.168 & 0.236 & 0.305 & 0.494 & 0.891 & 0.973 & 0.996 \\
DeepAR-MinT  & 0.483 & 0.137 & 0.166 & 0.226 & 0.284 & 0.428 & 0.842 & 0.946 & 0.975 \\
DeepAR-ERM   & 0.487 & 0.133 & 0.144 & 0.229 & 0.286 & 0.441 & 0.885 & 0.941 & 0.936 \\
HierE2E      & 0.438 & 0.119 & 0.142 & 0.205 & 0.244 & 0.428 & 0.914 & 0.952 & 0.983 \\
SHARQ        & 0.427 & 0.106 & 0.153 & 0.196 & 0.249 & 0.448 & 0.934 & 0.944 & 0.955 \\
        PEMBU-MINT   & 0.431 & 0.126 & 0.173 & 0.217 & 0.257 & 0.427 & 0.847 & 0.933 & 0.981 \\
        Dow Current & 0.443 & 0.117 & 0.194 & 0.227 & 0.294 & 0.491 & 0.921 & 0.932 & 0.995 \\
        ProfHiT      & \underline{0.421} & \textbf{0.101} & \underline{0.143} & \underline{0.194} & \underline{0.218} & \underline{0.399} & \underline{0.834} & \underline{0.873} & \underline{0.926} \\ \hline
        \model    & \textbf{0.405} & \underline{0.105} & \textbf{0.115} & \textbf{0.175} & \textbf{0.196} & \textbf{0.284} & \textbf{0.623} & \textbf{0.529} & \textbf{0.737}
    \end{tabular}
\end{table*}

%% file: Tables/DowCRPS.tex
\begin{table*}[h]
    \centering
    \caption{Normalized CRPS for Dow dataset. \model achieves the best (bold) or second-best (underline) performance across all levels of hierarchy.}
    \label{tab:dowcrps}
    \begin{tabular}{c|c|cccccccc}
        Model        & \textbf{Total} & L1    & L2    & L3    & L4    & L5    & L6    & L7    & L8    \\\hline
        Sparsity     &       & 0   & 0   & 3.69   & 15.61   & 21.05   & 32.72  & 36.49  & 40.93  \\\hline
        ARIMA        & 5.265 & 3.027 & 3.335 & 4.326 & 4.502 & 6.120 & 8.927 & 9.044 & 11.952 \\
RNN          & 0.783 & 0.308 & 0.313 & 0.358 & 0.393 & 0.683 & 0.858 & 1.116 & 0.957  \\
DeepAR       & 0.715 & 0.302 & 0.299 & 0.286 & 0.393 & 0.525 & 0.768 & 0.968 & 1.008  \\
DeepAR-MinT  & 0.218 & 0.285 & 0.276 & 0.264 & 0.342 & 0.473 & 0.705 & 0.899 & 0.912  \\
DeepAR-ERM   & 0.117 & 0.240 & 0.254 & 0.320 & 0.377 & 0.431 & 0.816 & 0.870 & 0.830  \\
HierE2E        & 0.161 & 0.088 & 0.284 & 0.298 & 0.416 & {0.513} & 0.871 & 0.865 & 0.881  \\
SHARQ   & 0.154 & 0.066 & \underline{0.073} & \textbf{0.074} & 0.326 & 0.494 & 0.873 & 0.939 & 0.849  \\
PEMBU-MinT & \underline{0.132} & \textbf{0.032} & 0.096 & 0.228 & \textbf{0.193} & \textbf{0.329} & \underline{0.473} & \underline{0.686} & 0.937  \\
ProfHiT      & 0.146 & \underline{0.033} & 0.099 & 0.203 & 0.342 & 0.548 & 0.663 & 0.754 & \underline{0.837}  \\\hline
\model    & \textbf{0.090} & \textbf{0.032} & \textbf{0.065} & \underline{0.154} & \underline{0.297} & \underline{0.480} & \textbf{0.453} & \textbf{0.658} & \textbf{0.746} 
    \end{tabular}
\end{table*}

%% file: Text/casestudy.tex
\subsubsection{Case Study: Demand Forecasting at Dow}

\noindent\textit{Background:}
At Dow, hierarchical time series models are developed to forecast product demand and raw material price to facilitate business planning. These models offer substantial value to the businesses by minimizing the cost to serve through improved planning and forecasting, thereby enhancing customer experience and relationships. Currently, forecasts are performed by applying Microsoft Azure Auto Machine Learning (AutoML), a cloud-based service that automates the selection and tuning of machine learning models. One of the main drawbacks of this approach is the restriction on the number of predictor variables that can be included in the model, resulting from poor model scalability. In addition, the relationships among different layers in the hierarchy are expected to provide useful insights on the product demand but are not accounted for in the model training (i.e., aggregated data at a higher level of granularity were provided for model training and inference and are disaggregated to lower levels based on proportions derived from historical data). Last but not the least, the lack of transparency, and uncertainty-driven risk assessment associated with the model performance and forecasts impose significant challenges to business decision-making processes.

    \begin{figure}[htb]
        \includegraphics[width=.7\linewidth]{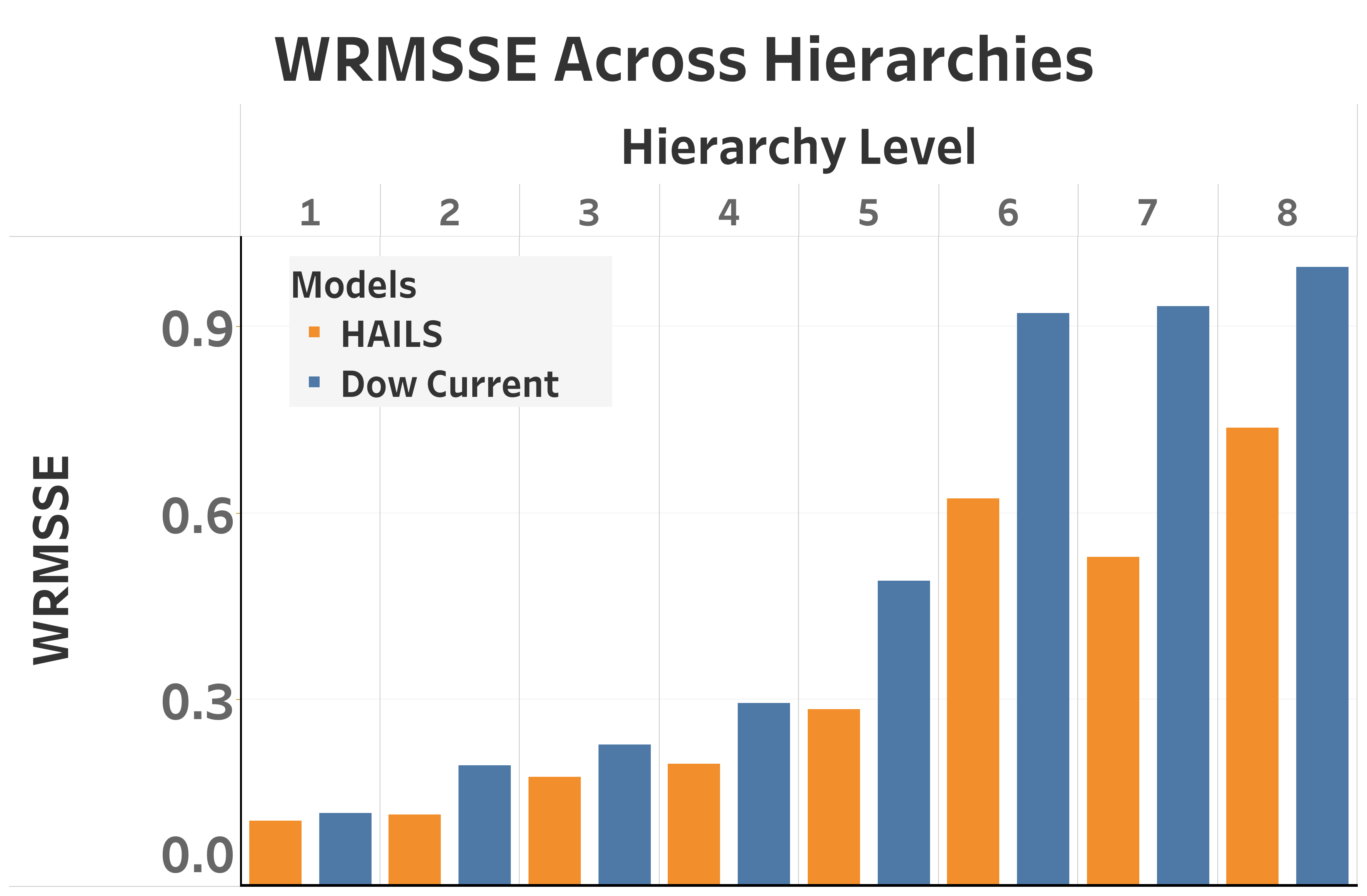}
        \caption{\model has significantly lower WRMSSE than Dow baseline across all levels of the hierarchy.}
        \label{fig:worldmap}
    \end{figure}

\noindent\textit{Impact:} We developed \model to overcome these challenges that are commonly existed in large-scale business planning.
\model alleviates these crucial challenges: First, it efficiently scales to predict the time-series across all levels of the hierarchy.
It additionally leverages DCRS to optimize for distributional consistency according to underlying hierarchical relationships.
Finally, being a state-of-art probabilistic model it provides reliable forecast distributions that are both accurate and have dependable uncertainty measures. These benefits allow HAILS to have a vastly lower RMSE across the entire forecast horizon.

\begin{figure}[h]
        \includegraphics[width=.8\linewidth]{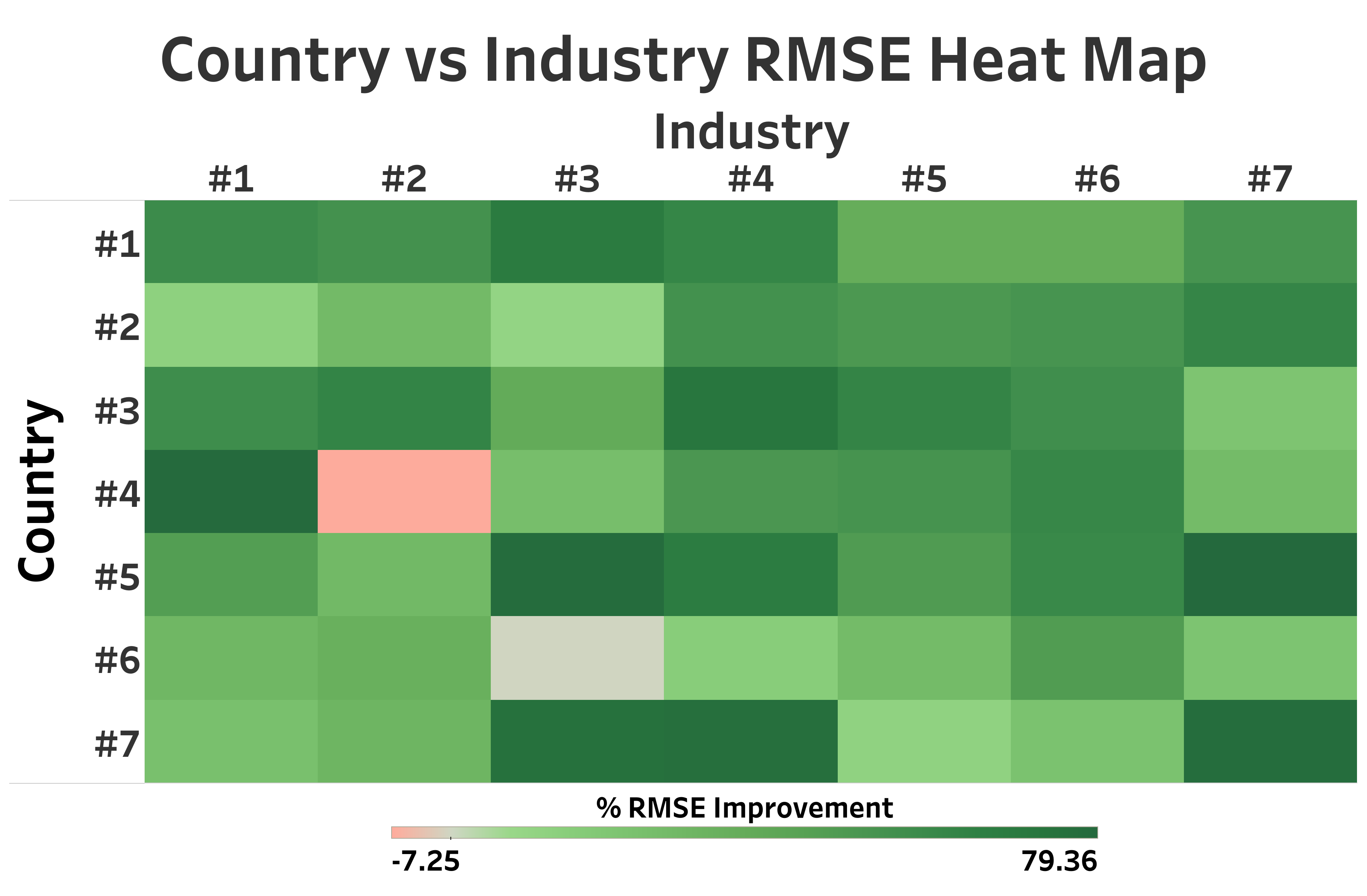}
        \caption{\model provides an average of 44.14\% improvement over Dow Model over top 7 industries and countries.}
        \label{fig:heatmaps}
    \end{figure}

\begin{figure}[h]
        \includegraphics[width=.8\linewidth]{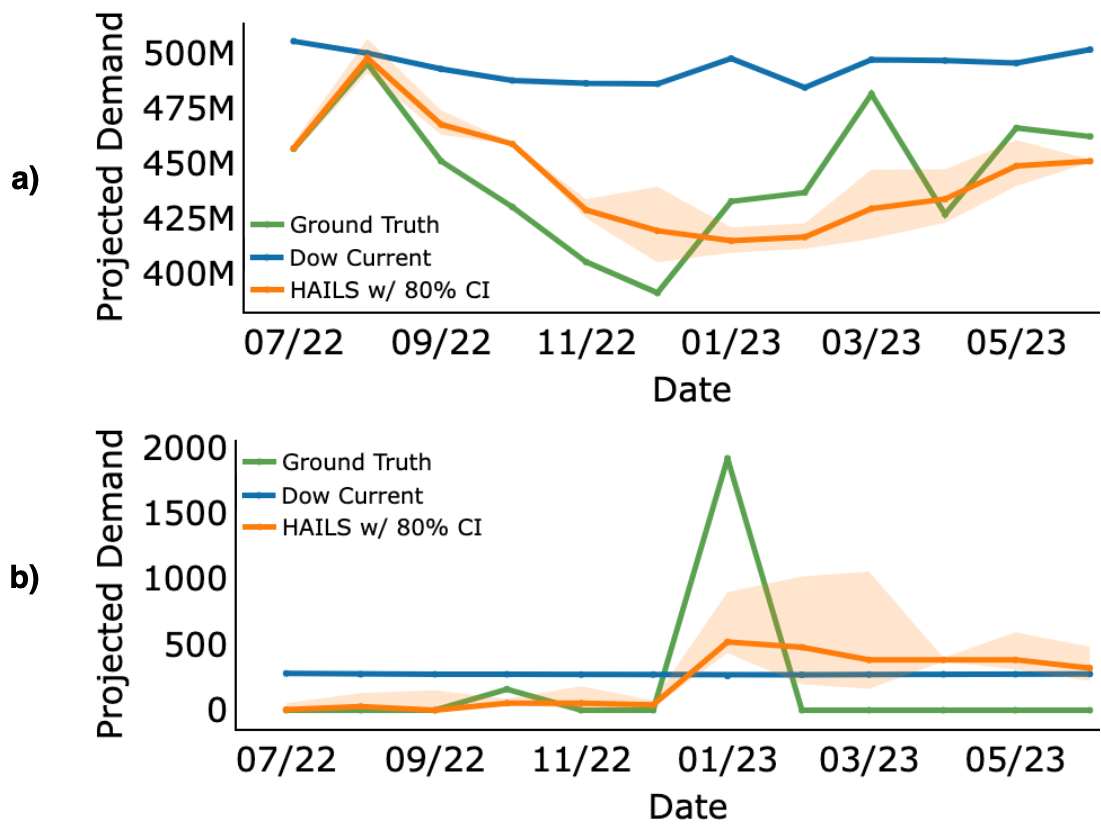}
        \caption{Examples of forecasts of \model and Dow baseline for (a) dense  and (b) sparse time-series. \model's forecasts are much more accurate with uncertainty bands close to the ground truth.}
        \label{fig:forecasts}
    \end{figure}

Historical demand is used as the criterion to identify the top countries and industries. This is based on the assumption that higher demand corresponds to higher value, and thus more potential for profit. By improving forecast accuracy for these segments, we can optimize our business planning and reduce costs.
We summarize the forecasting performance of \model, Dow's AutoML baseline and other baselines in Tables \ref{tab:dowrmsse}, \ref{tab:dowcrps}.
\model outperforms the previous baseline used by Dow by over 8.5\% overall in RMSSE with an average improvement of 26\% for the last three layers which have over 10\% of the values zeroes (Fig. \ref{fig:worldmap}). 
Similarly, \model's CRPS score is 30\% better than the best baseline models with over 23\% better in the last 3 sparser levels of the hierarchy.
\begin{figure}[h]
        \includegraphics[width=0.8\linewidth]{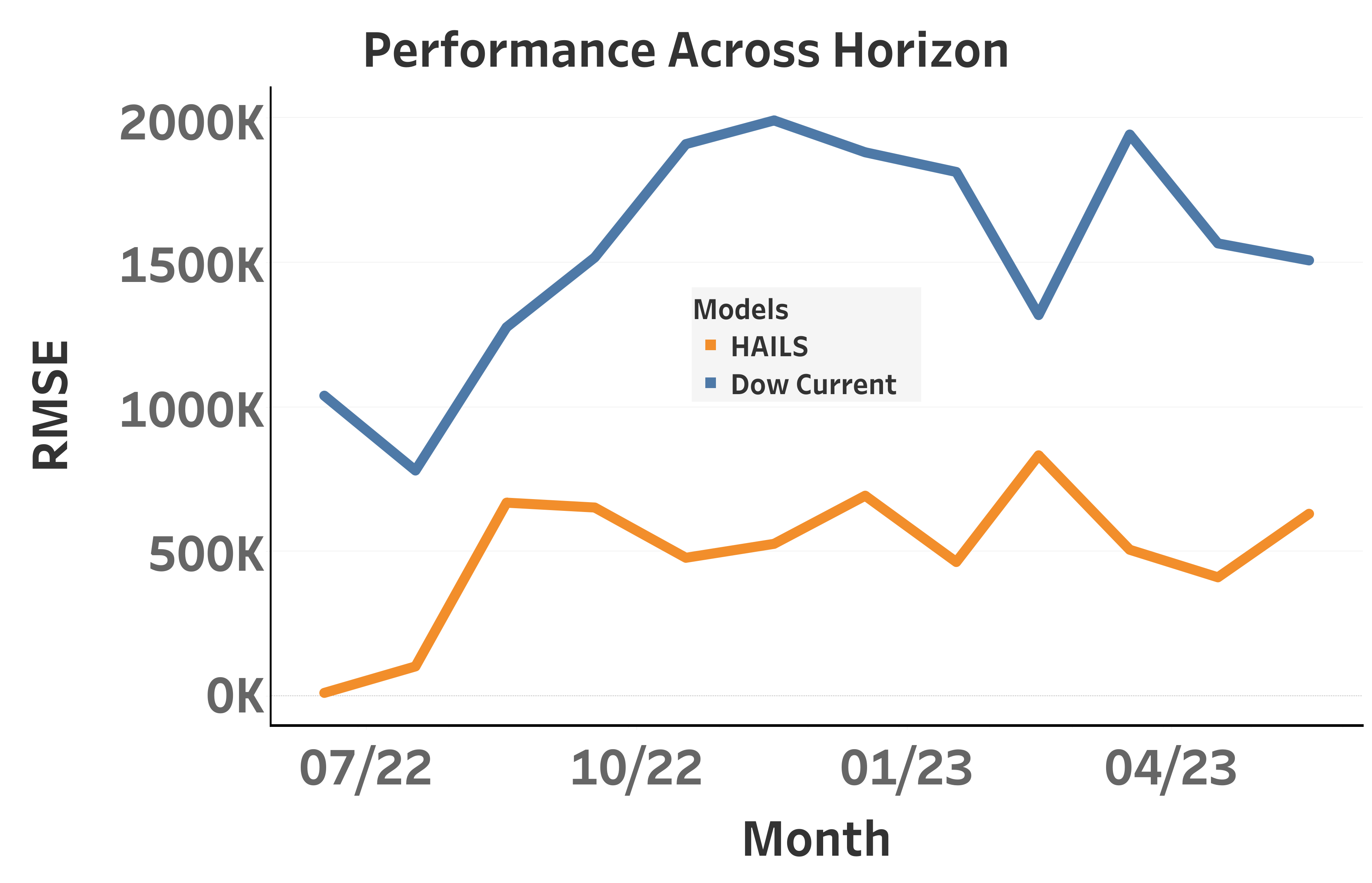}
        \caption{RMSE of \model is consistently lower than Dow baseline across the forecast horizon.}
        \label{fig:consistent}
    \end{figure}
The improvement in forecast performance is seen consistently during
testing across the year (Fig. \ref{fig:consistent}).
We also observe that \model is 70\% faster to train than the next best model.
We also observe 44.14\% average improvement in performance for forecasts in  the top
 seven countries and industries identified by magnitude of past demand (Fig.~\ref{fig:heatmaps}). 
We also visualize few examples forecasts.
We also observe that the confidence intervals of the forecasts
closely follow the ground truth compared the the Dow baseline (Fig.~\ref{fig:forecasts}).


\input{Tables/Eff.tex}
\subsubsection{Efficiency}

\model leverages DCSR and Poisson selection to model sparse time-series as Poisson distribution. 
Moreover, we also leverage pre-training to improve the convergence and training efficiency of the model since \model typically
takes lesser epochs to achieve state-of-the-art performance.

We measure the total training time in hours for the model and baselines in Table \ref{fig:enter-label}. 
We run the code on workstation with Intel Xeon CPU with 64 cores, 128 GB RAM and a Nvidia V100 GPU with 32GB VRAM.
\model is more efficient than end-to-end neural models like SHARQ, HierE2E and \profhit, finishing training in less than 42\% of the total time of the second best baseline for most of the datasets. This is due to effective modeling of sparse time series as well as asynchronous updates of model weights.

%% file: Tables/Eff.tex
\begin{figure}[htb]
    \centering
    \includegraphics[width=.9\linewidth]{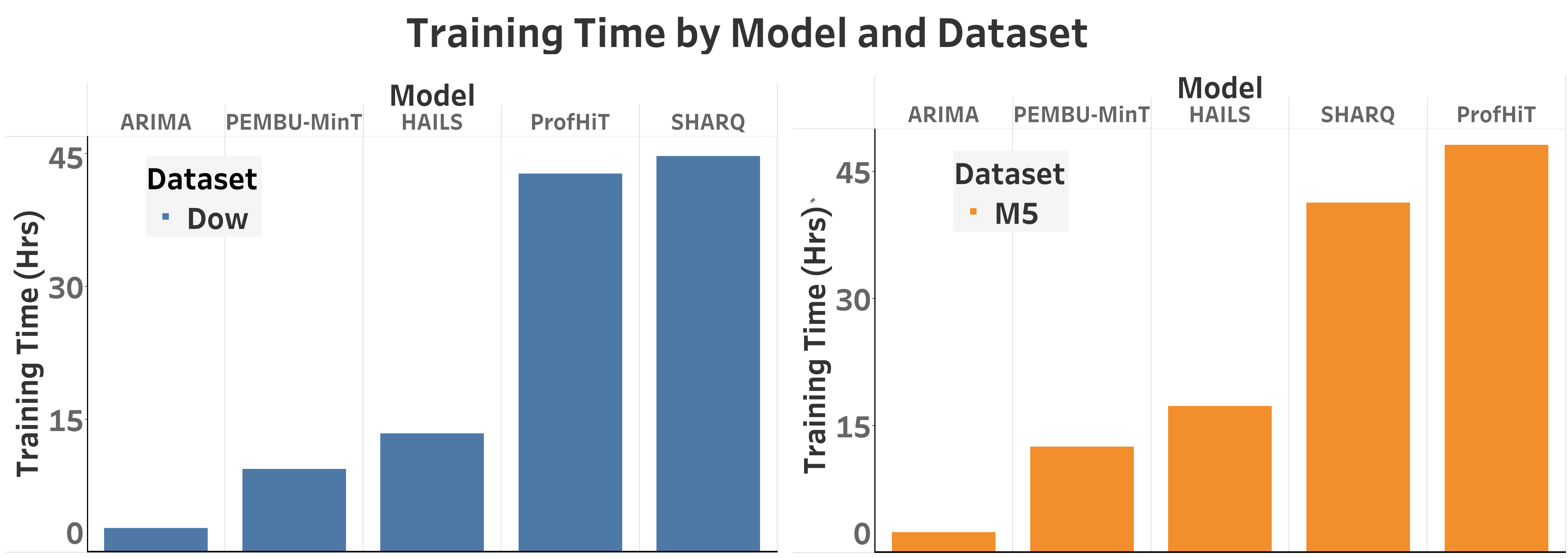}
    \caption{\model takes significantly less training time than state-of-art neural baselines like \profhit and SHARQ.
    }
    \label{fig:enter-label}
\end{figure}

%% file: Text/conclusion.tex
\section{Conclusion}
\model is designed to solve challenges motivated by our experience dealing with real-world large scale demand forecasting problem: scalability and modeling sparse time-series across the hierarchy.
 \model improves on \profhit to support sparse time-series at lower levels of the hierarchy, an important property of real-world demand forecasting scenario that enables it to perform 8-30\% better than previous best baselines with consistent performance across all levels of the hierarchy. \model also outperformed the baselines by over 20\% in the sparse layers of the hierarchy. Our model design and training enables \model to train up to three times more efficiently than similarly sized state-of-art models enabling effective and accurate real-time forecasting. Our model was successfully applied to a real-world application of demand forecasting in one of the world's largest chemical companies and yielded significantly superior performance across the hierarchy.
This enables significant reductions in cost due to manufacturing planning, inventory management and fulfillment scheduling.

 There are other deployment challenges for \model that include data collection, data cleaning, choosing the right hierarchy, explainability and deployment. Collecting reliable data across the hierarchy in a large corporation is complicated by the number of systems, businesses and geographical areas and various product units of measure that need to be standardized. 
 Therefore, building systems that can understand and leverage data quality information to improve the robustness of the forecasts is an important problem~\cite{kamarthi2021back2future}.
 Another important challenge is providing interpretability as block-box neural models are not readily accepted in the business process. 
 Developing reliable interpretability methods for hierarchical forecasting is essential for successful deployment. Additionally, 
 the hierarchy structures may change due to reasons such as reclassifications from one business grouping to another, addition or deletion of products, etc.
 While we can recalculate the time-series values of the past for new hierarchy, deriving information from a dynamic hierarchy structure is a novel research direction. 

 \section*{Acknowledgements}
 This paper was supported in part by The Dow Chemical Company, the NSF (Expeditions CCF-1918770, CAREER IIS-2028586, Medium IIS-1955883, Medium IIS-2106961, PIPP CCF-2200269), CDC MInD program, Meta faculty gifts, and funds/computing resources from Georgia Tech.

%% file: main.bbl

\begin{thebibliography}{26}


\ifx \showCODEN    \undefined \def \showCODEN     #1{\unskip}     \fi
\ifx \showDOI      \undefined \def \showDOI       #1{#1}\fi
\ifx \showISBNx    \undefined \def \showISBNx     #1{\unskip}     \fi
\ifx \showISBNxiii \undefined \def \showISBNxiii  #1{\unskip}     \fi
\ifx \showISSN     \undefined \def \showISSN      #1{\unskip}     \fi
\ifx \showLCCN     \undefined \def \showLCCN      #1{\unskip}     \fi
\ifx \shownote     \undefined \def \shownote      #1{#1}          \fi
\ifx \showarticletitle \undefined \def \showarticletitle #1{#1}   \fi
\ifx \showURL      \undefined \def \showURL       {\relax}        \fi
\providecommand\bibfield[2]{#2}
\providecommand\bibinfo[2]{#2}
\providecommand\natexlab[1]{#1}
\providecommand\showeprint[2][]{arXiv:#2}

\bibitem[\protect\citeauthoryear{Athanasopoulos, Gamakumara, Panagiotelis, Hyndman, and Affan}{Athanasopoulos et~al\mbox{.}}{2020}]%
        {athanasopoulos2020hierarchical}
\bibfield{author}{\bibinfo{person}{George Athanasopoulos}, \bibinfo{person}{Puwasala Gamakumara}, \bibinfo{person}{Anastasios Panagiotelis}, \bibinfo{person}{Rob~J Hyndman}, {and} \bibinfo{person}{Mohamed Affan}.} \bibinfo{year}{2020}\natexlab{}.
\newblock \showarticletitle{Hierarchical forecasting}.
\newblock \bibinfo{journal}{\emph{Macroeconomic forecasting in the era of big data: Theory and practice}} (\bibinfo{year}{2020}), \bibinfo{pages}{689--719}.
\newblock


\bibitem[\protect\citeauthoryear{Ben~Taieb and Koo}{Ben~Taieb and Koo}{2019}]%
        {ben2019regularized}
\bibfield{author}{\bibinfo{person}{Souhaib Ben~Taieb} {and} \bibinfo{person}{Bonsoo Koo}.} \bibinfo{year}{2019}\natexlab{}.
\newblock \showarticletitle{Regularized regression for hierarchical forecasting without unbiasedness conditions}. In \bibinfo{booktitle}{\emph{Proceedings of the 25th ACM SIGKDD International Conference on Knowledge Discovery \& Data Mining}}. \bibinfo{pages}{1337--1347}.
\newblock


\bibitem[\protect\citeauthoryear{B{\"o}se, Flunkert, Gasthaus, Januschowski, Lange, Salinas, Schelter, Seeger, and Wang}{B{\"o}se et~al\mbox{.}}{2017}]%
        {bose2017probabilistic}
\bibfield{author}{\bibinfo{person}{Joos-Hendrik B{\"o}se}, \bibinfo{person}{Valentin Flunkert}, \bibinfo{person}{Jan Gasthaus}, \bibinfo{person}{Tim Januschowski}, \bibinfo{person}{Dustin Lange}, \bibinfo{person}{David Salinas}, \bibinfo{person}{Sebastian Schelter}, \bibinfo{person}{Matthias Seeger}, {and} \bibinfo{person}{Yuyang Wang}.} \bibinfo{year}{2017}\natexlab{}.
\newblock \showarticletitle{Probabilistic demand forecasting at scale}.
\newblock \bibinfo{journal}{\emph{Proceedings of the VLDB Endowment}} \bibinfo{volume}{10}, \bibinfo{number}{12} (\bibinfo{year}{2017}), \bibinfo{pages}{1694--1705}.
\newblock


\bibitem[\protect\citeauthoryear{Chung, Gulcehre, Cho, and Bengio}{Chung et~al\mbox{.}}{2014}]%
        {chung2014empirical}
\bibfield{author}{\bibinfo{person}{Junyoung Chung}, \bibinfo{person}{Caglar Gulcehre}, \bibinfo{person}{KyungHyun Cho}, {and} \bibinfo{person}{Yoshua Bengio}.} \bibinfo{year}{2014}\natexlab{}.
\newblock \showarticletitle{Empirical evaluation of gated recurrent neural networks on sequence modeling}.
\newblock \bibinfo{journal}{\emph{arXiv preprint arXiv:1412.3555}} (\bibinfo{year}{2014}).
\newblock


\bibitem[\protect\citeauthoryear{Gal and Ghahramani}{Gal and Ghahramani}{2016}]%
        {gal2016dropout}
\bibfield{author}{\bibinfo{person}{Yarin Gal} {and} \bibinfo{person}{Zoubin Ghahramani}.} \bibinfo{year}{2016}\natexlab{}.
\newblock \showarticletitle{Dropout as a bayesian approximation: Representing model uncertainty in deep learning}. In \bibinfo{booktitle}{\emph{international conference on machine learning}}. PMLR, \bibinfo{pages}{1050--1059}.
\newblock


\bibitem[\protect\citeauthoryear{Han, Dasgupta, and Ghosh}{Han et~al\mbox{.}}{2021}]%
        {han2021simultaneously}
\bibfield{author}{\bibinfo{person}{Xing Han}, \bibinfo{person}{Sambarta Dasgupta}, {and} \bibinfo{person}{Joydeep Ghosh}.} \bibinfo{year}{2021}\natexlab{}.
\newblock \showarticletitle{Simultaneously Reconciled Quantile Forecasting of Hierarchically Related Time Series}. In \bibinfo{booktitle}{\emph{International Conference on Artificial Intelligence and Statistics}}. PMLR, \bibinfo{pages}{190--198}.
\newblock


\bibitem[\protect\citeauthoryear{Hinton and Salakhutdinov}{Hinton and Salakhutdinov}{2006}]%
        {hinton2006reducing}
\bibfield{author}{\bibinfo{person}{Geoffrey~E Hinton} {and} \bibinfo{person}{Ruslan~R Salakhutdinov}.} \bibinfo{year}{2006}\natexlab{}.
\newblock \showarticletitle{Reducing the dimensionality of data with neural networks}.
\newblock \bibinfo{journal}{\emph{science}} \bibinfo{volume}{313}, \bibinfo{number}{5786} (\bibinfo{year}{2006}), \bibinfo{pages}{504--507}.
\newblock


\bibitem[\protect\citeauthoryear{Hyndman, Ahmed, Athanasopoulos, and Shang}{Hyndman et~al\mbox{.}}{2011}]%
        {hyndman2011optimal}
\bibfield{author}{\bibinfo{person}{Rob~J Hyndman}, \bibinfo{person}{Roman~A Ahmed}, \bibinfo{person}{George Athanasopoulos}, {and} \bibinfo{person}{Han~Lin Shang}.} \bibinfo{year}{2011}\natexlab{}.
\newblock \showarticletitle{Optimal combination forecasts for hierarchical time series}.
\newblock \bibinfo{journal}{\emph{Computational statistics \& data analysis}} \bibinfo{volume}{55}, \bibinfo{number}{9} (\bibinfo{year}{2011}), \bibinfo{pages}{2579--2589}.
\newblock


\bibitem[\protect\citeauthoryear{Hyndman and Athanasopoulos}{Hyndman and Athanasopoulos}{2018}]%
        {hyndman2018forecasting}
\bibfield{author}{\bibinfo{person}{Rob~J Hyndman} {and} \bibinfo{person}{George Athanasopoulos}.} \bibinfo{year}{2018}\natexlab{}.
\newblock \bibinfo{booktitle}{\emph{Forecasting: principles and practice}}.
\newblock \bibinfo{publisher}{OTexts}.
\newblock


\bibitem[\protect\citeauthoryear{Jati, Ekambaram, Pal, Quanz, Gifford, Harsha, Siegel, Mukherjee, and Narayanaswami}{Jati et~al\mbox{.}}{2023}]%
        {jati2023hierarchical}
\bibfield{author}{\bibinfo{person}{Arindam Jati}, \bibinfo{person}{Vijay Ekambaram}, \bibinfo{person}{Shaonli Pal}, \bibinfo{person}{Brian Quanz}, \bibinfo{person}{Wesley~M Gifford}, \bibinfo{person}{Pavithra Harsha}, \bibinfo{person}{Stuart Siegel}, \bibinfo{person}{Sumanta Mukherjee}, {and} \bibinfo{person}{Chandra Narayanaswami}.} \bibinfo{year}{2023}\natexlab{}.
\newblock \showarticletitle{Hierarchical Proxy Modeling for Improved HPO in Time Series Forecasting}. In \bibinfo{booktitle}{\emph{Proceedings of the 29th ACM SIGKDD Conference on Knowledge Discovery and Data Mining}}. \bibinfo{pages}{891--900}.
\newblock


\bibitem[\protect\citeauthoryear{Kamarthi, Kong, Rodriguez, Zhang, and Prakash}{Kamarthi et~al\mbox{.}}{2021}]%
        {kamarthi2021doubt}
\bibfield{author}{\bibinfo{person}{Harshavardhan Kamarthi}, \bibinfo{person}{Lingkai Kong}, \bibinfo{person}{Alexander Rodriguez}, \bibinfo{person}{Chao Zhang}, {and} \bibinfo{person}{B~Aditya Prakash}.} \bibinfo{year}{2021}\natexlab{}.
\newblock \showarticletitle{When in Doubt: Neural Non-Parametric Uncertainty Quantification for Epidemic Forecasting}.
\newblock \bibinfo{journal}{\emph{Thirty-fifth Conference on Neural Information Processing Systems}} (\bibinfo{year}{2021}).
\newblock


\bibitem[\protect\citeauthoryear{Kamarthi, Kong, Rodriguez, Zhang, and Prakash}{Kamarthi et~al\mbox{.}}{2022a}]%
        {kamarthi2021camul}
\bibfield{author}{\bibinfo{person}{Harshavardhan Kamarthi}, \bibinfo{person}{Lingkai Kong}, \bibinfo{person}{Alexander Rodriguez}, \bibinfo{person}{Chao Zhang}, {and} \bibinfo{person}{B~Aditya Prakash}.} \bibinfo{year}{2022}\natexlab{a}.
\newblock \showarticletitle{CAMul: Calibrated and Accurate Multi-view Time-Series Forecasting}.
\newblock \bibinfo{journal}{\emph{ACM The Web Conference (WWW)}} (\bibinfo{year}{2022}).
\newblock


\bibitem[\protect\citeauthoryear{Kamarthi, Kong, Rodr{\'\i}guez, Zhang, and Prakash}{Kamarthi et~al\mbox{.}}{2023}]%
        {kamarthi2023rigidity}
\bibfield{author}{\bibinfo{person}{Harshavardhan Kamarthi}, \bibinfo{person}{Lingkai Kong}, \bibinfo{person}{Alexander Rodr{\'\i}guez}, \bibinfo{person}{Chao Zhang}, {and} \bibinfo{person}{B~Aditya Prakash}.} \bibinfo{year}{2023}\natexlab{}.
\newblock \showarticletitle{When Rigidity Hurts: Soft Consistency Regularization for Probabilistic Hierarchical Time Series Forecasting}. In \bibinfo{booktitle}{\emph{Proceedings of the 29th ACM SIGKDD Conference on Knowledge Discovery and Data Mining}}. \bibinfo{pages}{1057--1072}.
\newblock


\bibitem[\protect\citeauthoryear{Kamarthi, Rodr{\'\i}guez, and Prakash}{Kamarthi et~al\mbox{.}}{2022b}]%
        {kamarthi2021back2future}
\bibfield{author}{\bibinfo{person}{Harshavardhan Kamarthi}, \bibinfo{person}{Alexander Rodr{\'\i}guez}, {and} \bibinfo{person}{B~Aditya Prakash}.} \bibinfo{year}{2022}\natexlab{b}.
\newblock \showarticletitle{Back2future: Leveraging backfill dynamics for improving real-time predictions in future}.
\newblock \bibinfo{journal}{\emph{ICLR}} (\bibinfo{year}{2022}).
\newblock


\bibitem[\protect\citeauthoryear{Lesch and Jeske}{Lesch and Jeske}{2009}]%
        {lesch2009some}
\bibfield{author}{\bibinfo{person}{Scott~M Lesch} {and} \bibinfo{person}{Daniel~R Jeske}.} \bibinfo{year}{2009}\natexlab{}.
\newblock \showarticletitle{Some suggestions for teaching about normal approximations to poisson and binomial distribution functions}.
\newblock \bibinfo{journal}{\emph{The American Statistician}} \bibinfo{volume}{63}, \bibinfo{number}{3} (\bibinfo{year}{2009}), \bibinfo{pages}{274--277}.
\newblock


\bibitem[\protect\citeauthoryear{Louizos, Shi, Schutte, and Welling}{Louizos et~al\mbox{.}}{2019}]%
        {louizos2019functional}
\bibfield{author}{\bibinfo{person}{Christos Louizos}, \bibinfo{person}{Xiahan Shi}, \bibinfo{person}{Klamer Schutte}, {and} \bibinfo{person}{Max Welling}.} \bibinfo{year}{2019}\natexlab{}.
\newblock \showarticletitle{The functional neural process}.
\newblock \bibinfo{journal}{\emph{arXiv preprint arXiv:1906.08324}} (\bibinfo{year}{2019}).
\newblock


\bibitem[\protect\citeauthoryear{Makridakis and Hibon}{Makridakis and Hibon}{1997}]%
        {makridakis1997arma}
\bibfield{author}{\bibinfo{person}{Spyros Makridakis} {and} \bibinfo{person}{Michele Hibon}.} \bibinfo{year}{1997}\natexlab{}.
\newblock \showarticletitle{ARMA models and the Box--Jenkins methodology}.
\newblock \bibinfo{journal}{\emph{Journal of forecasting}} \bibinfo{volume}{16}, \bibinfo{number}{3} (\bibinfo{year}{1997}), \bibinfo{pages}{147--163}.
\newblock


\bibitem[\protect\citeauthoryear{Makridakis, Spiliotis, and Assimakopoulos}{Makridakis et~al\mbox{.}}{2022}]%
        {makridakis2022m5}
\bibfield{author}{\bibinfo{person}{Spyros Makridakis}, \bibinfo{person}{Evangelos Spiliotis}, {and} \bibinfo{person}{Vassilios Assimakopoulos}.} \bibinfo{year}{2022}\natexlab{}.
\newblock \showarticletitle{M5 accuracy competition: Results, findings, and conclusions}.
\newblock \bibinfo{journal}{\emph{International Journal of Forecasting}} \bibinfo{volume}{38}, \bibinfo{number}{4} (\bibinfo{year}{2022}), \bibinfo{pages}{1346--1364}.
\newblock


\bibitem[\protect\citeauthoryear{Rangapuram, Werner, Benidis, Mercado, Gasthaus, and Januschowski}{Rangapuram et~al\mbox{.}}{2021}]%
        {rangapuram2021end}
\bibfield{author}{\bibinfo{person}{Syama~Sundar Rangapuram}, \bibinfo{person}{Lucien~D Werner}, \bibinfo{person}{Konstantinos Benidis}, \bibinfo{person}{Pedro Mercado}, \bibinfo{person}{Jan Gasthaus}, {and} \bibinfo{person}{Tim Januschowski}.} \bibinfo{year}{2021}\natexlab{}.
\newblock \showarticletitle{End-to-End Learning of Coherent Probabilistic Forecasts for Hierarchical Time Series}. In \bibinfo{booktitle}{\emph{International Conference on Machine Learning}}. PMLR, \bibinfo{pages}{8832--8843}.
\newblock


\bibitem[\protect\citeauthoryear{Reich, Brooks, Fox, Kandula, McGowan, Moore, Osthus, Ray, Tushar, Yamana, et~al\mbox{.}}{Reich et~al\mbox{.}}{2019}]%
        {reich2019collaborative}
\bibfield{author}{\bibinfo{person}{Nicholas~G Reich}, \bibinfo{person}{Logan~C Brooks}, \bibinfo{person}{Spencer~J Fox}, \bibinfo{person}{Sasikiran Kandula}, \bibinfo{person}{Craig~J McGowan}, \bibinfo{person}{Evan Moore}, \bibinfo{person}{Dave Osthus}, \bibinfo{person}{Evan~L Ray}, \bibinfo{person}{Abhinav Tushar}, \bibinfo{person}{Teresa~K Yamana}, {et~al\mbox{.}}} \bibinfo{year}{2019}\natexlab{}.
\newblock \showarticletitle{A collaborative multiyear, multimodel assessment of seasonal influenza forecasting in the United States}.
\newblock \bibinfo{journal}{\emph{Proceedings of the National Academy of Sciences}} \bibinfo{volume}{116}, \bibinfo{number}{8} (\bibinfo{year}{2019}), \bibinfo{pages}{3146--3154}.
\newblock


\bibitem[\protect\citeauthoryear{Salinas, Flunkert, Gasthaus, and Januschowski}{Salinas et~al\mbox{.}}{2020}]%
        {salinas2020deepar}
\bibfield{author}{\bibinfo{person}{David Salinas}, \bibinfo{person}{Valentin Flunkert}, \bibinfo{person}{Jan Gasthaus}, {and} \bibinfo{person}{Tim Januschowski}.} \bibinfo{year}{2020}\natexlab{}.
\newblock \showarticletitle{DeepAR: Probabilistic forecasting with autoregressive recurrent networks}.
\newblock \bibinfo{journal}{\emph{International Journal of Forecasting}} \bibinfo{volume}{36}, \bibinfo{number}{3} (\bibinfo{year}{2020}), \bibinfo{pages}{1181--1191}.
\newblock


\bibitem[\protect\citeauthoryear{Syntetos and Boylan}{Syntetos and Boylan}{2005}]%
        {syntetos2005accuracy}
\bibfield{author}{\bibinfo{person}{Aris~A Syntetos} {and} \bibinfo{person}{John~E Boylan}.} \bibinfo{year}{2005}\natexlab{}.
\newblock \showarticletitle{The accuracy of intermittent demand estimates}.
\newblock \bibinfo{journal}{\emph{International Journal of forecasting}} \bibinfo{volume}{21}, \bibinfo{number}{2} (\bibinfo{year}{2005}), \bibinfo{pages}{303--314}.
\newblock


\bibitem[\protect\citeauthoryear{Taieb, Taylor, and Hyndman}{Taieb et~al\mbox{.}}{2017}]%
        {taieb2017coherent}
\bibfield{author}{\bibinfo{person}{Souhaib~Ben Taieb}, \bibinfo{person}{James~W Taylor}, {and} \bibinfo{person}{Rob~J Hyndman}.} \bibinfo{year}{2017}\natexlab{}.
\newblock \showarticletitle{Coherent probabilistic forecasts for hierarchical time series}. In \bibinfo{booktitle}{\emph{International Conference on Machine Learning}}. PMLR, \bibinfo{pages}{3348--3357}.
\newblock


\bibitem[\protect\citeauthoryear{T{\"u}rkmen, Januschowski, Wang, and Cemgil}{T{\"u}rkmen et~al\mbox{.}}{2021}]%
        {turkmen2021forecasting}
\bibfield{author}{\bibinfo{person}{Ali~Caner T{\"u}rkmen}, \bibinfo{person}{Tim Januschowski}, \bibinfo{person}{Yuyang Wang}, {and} \bibinfo{person}{Ali~Taylan Cemgil}.} \bibinfo{year}{2021}\natexlab{}.
\newblock \showarticletitle{Forecasting intermittent and sparse time series: A unified probabilistic framework via deep renewal processes}.
\newblock \bibinfo{journal}{\emph{Plos one}} \bibinfo{volume}{16}, \bibinfo{number}{11} (\bibinfo{year}{2021}), \bibinfo{pages}{e0259764}.
\newblock


\bibitem[\protect\citeauthoryear{Wickramasuriya}{Wickramasuriya}{2021}]%
        {wickramasuriya2021probabilistic}
\bibfield{author}{\bibinfo{person}{Shanika~L Wickramasuriya}.} \bibinfo{year}{2021}\natexlab{}.
\newblock \showarticletitle{Probabilistic forecast reconciliation under the Gaussian framework}.
\newblock \bibinfo{journal}{\emph{arXiv preprint arXiv:2103.11128}} (\bibinfo{year}{2021}).
\newblock


\bibitem[\protect\citeauthoryear{Wickramasuriya, Athanasopoulos, and Hyndman}{Wickramasuriya et~al\mbox{.}}{2019}]%
        {wickramasuriya2019optimal}
\bibfield{author}{\bibinfo{person}{Shanika~L Wickramasuriya}, \bibinfo{person}{George Athanasopoulos}, {and} \bibinfo{person}{Rob~J Hyndman}.} \bibinfo{year}{2019}\natexlab{}.
\newblock \showarticletitle{Optimal forecast reconciliation for hierarchical and grouped time series through trace minimization}.
\newblock \bibinfo{journal}{\emph{J. Amer. Statist. Assoc.}} \bibinfo{volume}{114}, \bibinfo{number}{526} (\bibinfo{year}{2019}), \bibinfo{pages}{804--819}.
\newblock


\end{thebibliography}
